\documentclass[twocolumn,3p]{elsarticle}

\usepackage{hyperref}

\journal{Expert Systems with Applications}

\biboptions{authoryear}

\usepackage[english]{babel}
\usepackage[T1]{fontenc}
\usepackage{graphicx}
\usepackage{dblfloatfix}
\graphicspath{{figures/}}
\usepackage{epstopdf} 

\usepackage{url}

\usepackage[]{algorithm}
\usepackage[noend]{algpseudocode}
\algnewcommand\And{\textbf{and}}

\usepackage{multirow}
\usepackage{tabularx}
\newcolumntype{Y}{>{\raggedleft\let\newline\\\arraybackslash\hspace{0pt}}X}
\newcolumntype{Z}{>{\centering\let\newline\\\arraybackslash\hspace{0pt}}X}
\usepackage{booktabs}
\usepackage[para,online,flushleft]{threeparttable}
\usepackage{subfigure}

\usepackage{color,soul}
\usepackage{xargs}                      
\usepackage[colorinlistoftodos,prependcaption,textsize=tiny]{todonotes}
\newcommandx{\unsure}[2][1=]{\todo[linecolor=red,backgroundcolor=red!25,bordercolor=red,#1]{#2}}
\newcommandx{\change}[2][1=]{\todo[linecolor=blue,backgroundcolor=blue!25,bordercolor=blue,#1]{#2}}
\newcommandx{\info}[2][1=]{\todo[linecolor=OliveGreen,backgroundcolor=OliveGreen!25,bordercolor=OliveGreen,#1]{#2}}
\newcommandx{\improvement}[2][1=]{\todo[linecolor=Plum,backgroundcolor=Plum!25,bordercolor=Plum,#1]{#2}}
\newcommandx{\questiont}[2][1=]{\todo[linecolor=green,backgroundcolor=green!25,bordercolor=green,#1]{#2}}
\newcommandx{\thiswillnotshow}[2][1=]{\todo[disable,#1]{#2}}
\usepackage[utf8]{inputenc}

\usepackage{epstopdf}
\begin{document}

\title{Using Metaheuristics for the Location of Bicycle Stations}
\begin{frontmatter}

\author[uma]{C.~Cintrano\corref{cor1}}
\ead{cintrano@lcc.uma.es}
\author[uma]{F.~Chicano}
\ead{chicano@lcc.uma.es}
\author[uma]{E.~Alba}
\ead{eat@lcc.uma.es}
\cortext[cor1]{Corresponding author}
\address[uma]{Universidad de M\'alaga, E.T.S.I. Inform\'atica, Bulevar Louis Pasteur 35, 29071 Malaga, Spain}


\begin{abstract}
In this work, we solve the problem of finding the best locations to place stations for depositing/collecting shared bicycles. To do this, we model the problem as the \textit{p}-median problem, that is a major existing localization problem in optimization. The \textit{p}-median problem seeks to place a set of facilities (bicycle stations) in a way that minimizes the distance between a set of clients (citizens) and their closest facility (bike station).  

We have used a genetic algorithm, iterated local search, particle swarm optimization, simulated annealing, and variable neighbourhood search, to find the best locations for the bicycle stations and study their comparative advantages. We use irace to parameterize each algorithm automatically, to contribute with a methodology to fine-tune algorithms automatically. We have also studied different real data (distance and weights) from diverse open data sources from a real city, Malaga (Spain), hopefully leading to a final smart city application. We have compared our results with the implemented solution in Malaga. Finally, we have analyzed how we can use our proposal to improve the existing system in the city by adding more stations.
\end{abstract}


\end{frontmatter}

\section{Introduction}
\label{sec:introduction}
In recent years there has been an increase in the options for citizens to move around the city.
One of the reasons is the appearance of new transport models such as car-sharing applications, transportation network companies, electric cars, etc.
We will focus on bicycle-sharing systems, and we will use data from the public bicycle-sharing system provided by the municipality of Malaga, Spain, to contrast our hypothesis in a realistic scenario.
Bicycle-sharing systems present multiple problems to be optimized: how many bicycles to place in each station, the routes to transport bike lots from one station to another, load balancing at the stations, selecting where to place the stations, etc.
Most of them have been well studied in the scientific literature~\citep{Singhvi2015,Hu2014,Chira2014,Chen2018}. 
To support these systems, in this article, we will focus on the problem of allocating stations where users can pick up/deposit public bicycles.
The location of the stations is a vitally important issue for this kind of system but it is less studied than the other problems. 
Also, it has a direct impact on any future improvements to these systems.
These systems usually base their models only on topological aspects~\citep{Chen2015,Liu2015a,Park2017}. We propose a change of perspective, focusing on how the citizens (customers of the system) are distributed across the city and how they use the system, instead of how the city is shaped.
The services offered by a city to its citizens must be adapted to the needs of the inhabitants, rather than the users of these services (the citizens) having to adapt to them.

There are multiple problems in the scientific literature for locating resources~\citep{drezner2001facility}: Quadratic assignment, \textit{p}-center, \textit{p}-mean, etc.
One of the most popular location problems is the \textit{p}-median problem~\citep{daskin2015p}.
It consists of finding the location of a set of facilities in a way that minimizes the distance from a set of clients to their nearest facility.
This problem arises especially when deciding the location of the infrastructure of a city, a needed task in today's smart cities.
We will formulate the problem of allocating bicycle stations as a \textit{p}-median problem, and using open data we will find locations that suit the citizens of Malaga.
Due to our modelling as \textit{p}-median, we have a model as powerful as other existing in the scientific literature~\citep{Chen2015,Liu2015a,Park2017}, but that allows us to use a more exhaustive state-of-the-art.
For example, we can use metaheuristic algorithms to solve this problem since they have proved their usefulness in solving the \textit{p}-median problem.

This work is an extension of the conference paper by~\cite{Cintrano2018}.
The contributions of the present work are:
\begin{itemize}
    \item We have considered not only real population data, city maps, geographic locations of stations, but we have also included information on the use that citizens make of the current system (bicycle collections and deposits).
    \item We have used five different algorithms (instead of just one) to solve the problem: Genetic Algorithm~\citep{GA}, Iterated local search~\citep{ILS}, Particle Swarm Optimization~\citep{PSO}, Simulated Annealing~\citep{SA}, and Variable Neighbourhood Search~\citep{VNS}. 
    \item We have configured all these algorithms using irace~\citep{LOPEZIBANEZ201643}, in order to optimize the parameters for each of them. We have added more operators to modify the solutions than in the conference paper (see Section~\ref{sec:algorithms}), both general and specific for the different algorithms.
    \item We have not just been looking for a ranking of algorithm performances, but also for finding out which models better approach the current needs of users. We have also compared our results using statistical tests.
    \item We have carried out an in-depth analysis of the evolution of fitness during the execution of the algorithms and how we can improve the current system of Malaga by adding more stations. 
\end{itemize}


The rest of this article is organised as follows:
Section~\ref{sec:background} presents the formulation of the \textit{p}-median problem.
Section~\ref{sec:problem} models the problem of location of bicycle stations and the various realistic data used in this study.
Section~\ref{sec:algorithms} describes the selected optimisation algorithms.
Section~\ref{sec:hw} summarizes the computing platform used in the experimentation analysed in Section~\ref{sec:experimentation}.
We discuss related work in Section~\ref{sec:rw} and conclude in Section~\ref{sec:conclusions}.

\section{Background}
\label{sec:background}
The $p$-median problem is one of the most-studied NP-hard discrete location problems \citep{Dantrakul2014a,Megiddot1984}. It can be formulated as follows:
Given a set of customers $N$ and a set $L$ of $p$ facilities in $F$, the $p$-median problem seeks to allocate $p$ facilities in $F$ while minimising the weighted sum of the distances between the customers and their closest facility. 
Formally, the problem is defined as:
\begin{equation}
\min \sum_{i = 1}^{|N|} w_i \min_{j \in L}d_{ij},\label{eq:p-median}
\end{equation}
where $L \subseteq F$, with $|L| = p$ is the set of potential locations for the facilities, $w_i$ is the weight of the customer $i$, and $d_{ij}$ is the distance between customer $i$ and facility $j$. If we have $w_i = 1,\, i = 1, \ldots, |N|$, that is, all weights are one, we have the unweighted version of the problem.

In our study, the customers are the citizens and the facilities are the stations. 
Any street segment between two intersections can be a station location.
We selected this formulation of the problem for two reasons: 
(i) the model is easy to understand and implement, and 
(ii) the $p$-median problem is a classical location problem that has been well-studied in the scientific literature \citep{Dantrakul2014a}. From the allocation of bicycle stations perspective, the $p$-median may serve as a good model to identify interesting places and relevant distributions of the stations across an urban area.

There are other formulations to the location of bicycle stations problem~\citep{Chen2015,Liu2015a,Kloimullner2017}, requiring different information. 
The \textit{p}-median problem requires little information (only the distance matrix and the weights). Although this may seem like a limitation, it is relatively simple to add additional information, either by pre-processing the weights or distances, (e.g., by considering the slopes of the streets) or by adding terms in the formulation itself (e.g., by adding capacity information related to the slots in each bicycle parking site). 

\begin{figure*}[t]
\centering
\includegraphics[width=0.95\textwidth]{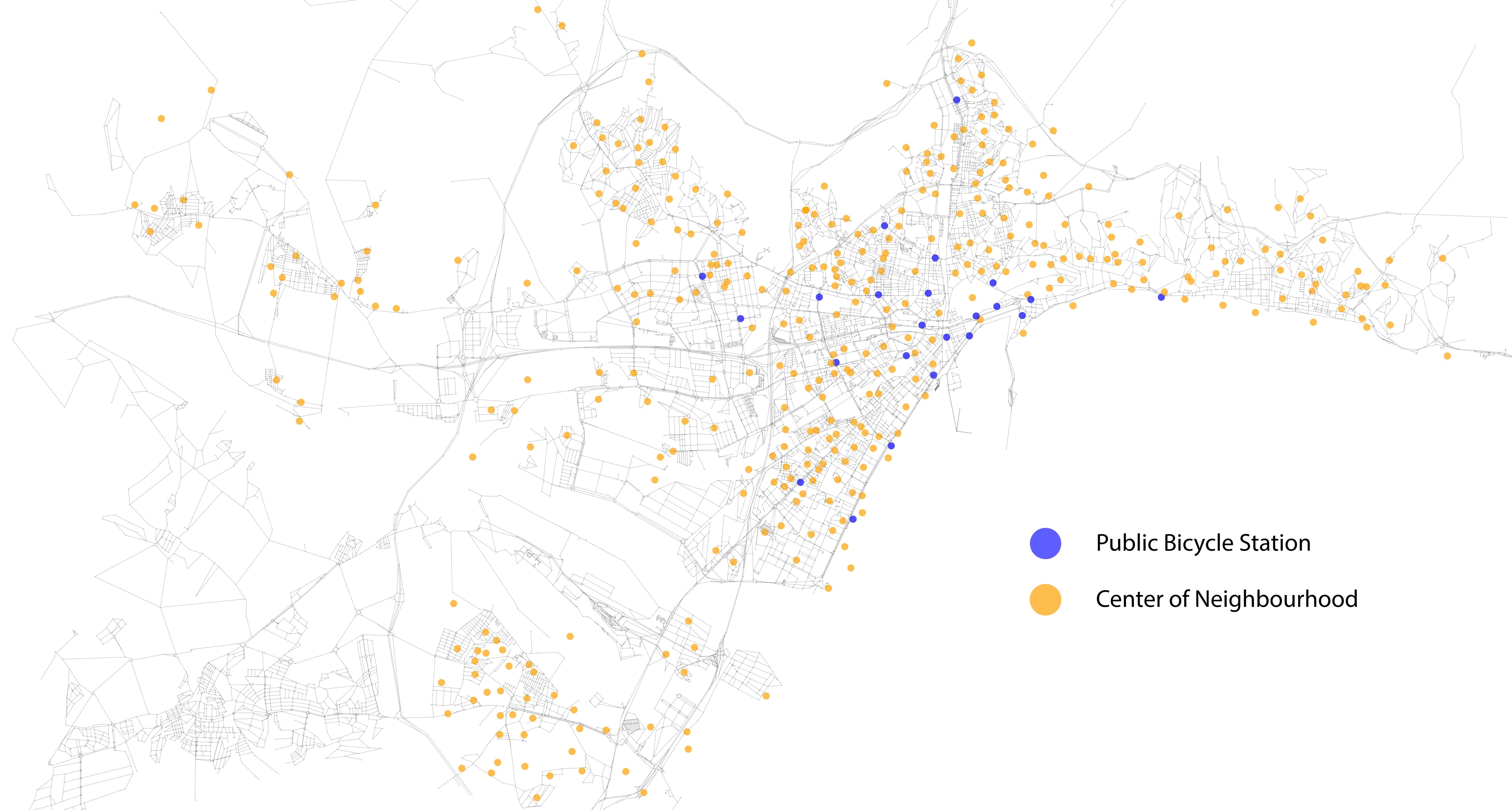}
    \vspace{-1em}
\caption{The centre of each neighbourhood (orange points) and current bicycle stations in the city of Malaga (blue points).}
\label{fig:malaga-city-graph}
\end{figure*}

\section{Problem Definition}
\label{sec:problem}
Our problem is to find the optimal positions for public bicycle stations so that citizens have to walk as little as possible to them.
Instead of using synthetic benchmark instances, we took advantage of many open data offered by the city council and specialised websites. In this way, we worked with a realistic instance as the city of Malaga.
We have decided to use this city as the target of our study for several reasons:
\begin{itemize}
\item It is a medium-sized European city with a variety of smart city initiatives. 
\item It has a wide variety of open data (see municipality open data website: \url{http://datosabiertos.malaga.eu/}), i.e., population census, filling status of bicycle stations, cartographic maps, etc.
\item It has already a shared bikes system with 23 public accessible stations, which will allow us to compare our proposed solutions using the $p$-median problem to the present situation.
\end{itemize}

In addition, the formulation of our problem as a $p$-median needs two sets of points: customers and facilities.
The customers are the citizens and the facilities are the bicycle stations.
We use the open data provided by the city council to have reliable information on both types of locations. 
The smallest group of citizens for whom we have location information is a neighbourhood. 
We have chosen the centres of the different neighbourhoods ($|N| = 363$) of the city as the most natural way of grouping the population (we excluded six neighbourhoods on the outskirts of Malaga).
The centre of each these neighbourhood is the position assigned to each customer in our problem.
Also, we selected all possible street segments as potential locations to the facilities.
In total, there are $|F| = 33,550$ potential locations.

Figure~\ref{fig:malaga-city-graph} shows the layout of the city, the centres of their neighborhood (orange points), and the 23 public bicycle stations (blue points).
We observe a greater number of stations in the central area of the city and only a few stations on the outskirts. This distribution does not have to be useful for the citizens since a large part of the population and points of interest (universities, sports centres, schools, etc.) are on the outskirts area. For this reason, a study carried out using intelligent techniques is useful not only when we need to install infrastructure for citizens, but also to assess the quality of existing solutions.

In our study, we have taken into account different values of distances and weights used in the \textit{p}-median problem.
In this way, we will be able to study how these data sets affect and with which data we obtain results that adapt better to the city.
We have used two types of distance $d$ between customers and stations: Straight-line Euclidean distance and the shortest path through the city streets (calculated using Dijkstra's algorithm), we refer to this last one as \textit{real distance}.


In addition, we have taken into account three types of weights $w_i, i \in N$, that a final application could consider to model the demand on each customer. These weights are:
\begin{itemize}
    \item Uniform: $w_i = 1$, is the basic option for solving the standard \textit{p}-median. It deals with the whole population fairly.
    \item Citizens: $w_i = c_i$, where $c_i$ is the number of residents in the $i$-th neighbourhood of the city. Neighbourhoods with more people should have a closer station.
    \item Demand: $w_i = p_i$, where $p_i$ is an estimate of demand in the $i$-th neighbourhood of the city. We have modeled this estimate based on the data of the use of the city's current public bicycle system. The open data about the stations allowed us to calculate the activity of each station as $act_j = mean(o_j)/mean(s_j)$, where $o_j$ are the number of slots occupied and $s_j$ the total number of slots in the $j$-th station. 
    We calculate the mean of $o_j$ and $s_j$ using the samples each one-minute intervals for a whole week (from 7th October 2018 to 13th October 2018) and we compute the value of $act_j$ in each station. Finally, we calculated the demand as $p_i = c_i \cdot act_{near_i}$, where $c_i$ is the number of inhabitants of the $i$-th neighbourhood and $act_{near_i}$ is the activity of the nearest station to the $i$-th neighbourhood.
\end{itemize}

\section{Algorithms}
\label{sec:algorithms}
In this article, we have used three metaheuristic algorithms based on trajectory search: Iterated Local Search (ILS), Simulated Annealing (SA), and Variable Neighbourhood Search (VNS); and two population-based algorithms: Genetic Algorithm (GA) and Particle Swarm Optimization (PSO). We will provide a brief description of all of them.

Genetic algorithm~\citep{GA} was originally presented by John Holland.
A basic pseudocode is showed in Algorithm~\ref{algorithm:ga}.
\begin{algorithm}
\caption{Genetic Algorithm}
\label{algorithm:ga}
\small
\begin{algorithmic}[1]
\State $pop \gets generatepopulaion()$
\State $i \gets 1$
\While{$i \leq iter$ $\And$ \ non stopping condition}
    \State $pop ^\prime \gets \emptyset$
    \For{$l \in \{1 ..\lambda\}$}
    	\State $parent_1, parent_2 \gets select(pop)$
    	\State $x \gets crossover(parent_1, parent_2)$
    	\State $x^\prime \gets mutation(x)$ 
    	\State $pop^\prime \gets x^\prime$
	\EndFor
	\State $pop \gets replacement(pop, pop^\prime)$
	\State $i \gets i + 1$
\EndWhile
\State \Return{pop}
\end{algorithmic}
\end{algorithm}
This kind of algorithms is inspired by natural evolution. 
In each iteration, the algorithm generates $\lambda$ new solutions (new population). A new solution is generated from several parent solutions (two parents in our case) selected from the previous population. The selected solutions are mixed (crossover) between them to generate a new one, which is probabilistically disturbed (mutated). At the end of an iteration, the new solutions replace others from the previous population following some kind of strategy. Finally, the algorithm returns the best solution found.  

Iterated local search was introduced by~\cite{ILS}. 
The pseudocode is presented in Algorithm~\ref{algorithm:ils}.
ILS performs a shake on the current (best) solution $x$ and applies a local search to get the nearest local optimal $x^\prime$. If $x^\prime$ improves on the best solution found so far $x$, it becomes the new best solution otherwise, it is discarded. 
Given the power of local searches in this type of localization problems, ILS has been used to solve the \textit{p}-median~\citep{ERDOGAN2016280,HO2015169}.

\begin{algorithm}
\caption{Iterated Local Search}
\label{algorithm:ils}
\small
\begin{algorithmic}[1]
\State $x \gets generation()$
\State $x \gets localsearch(x)$
\State $i \gets 1$
\While{$i \leq iter$ $\And$ \ non stopping condition}
	\State $x^\prime \gets shake(x, npert)$
	\State $x^\prime \gets localsearch(x^\prime)$
	\State $x \gets acceptation(x, x^\prime)$ // always elitist
	\State $i \gets i + 1$
\EndWhile
\State \Return{x}
\end{algorithmic}
\end{algorithm}

Particle swarm optimization was originally proposed by~\cite{PSO}.
Algorithm~\ref{algorithm:pso} shows a pseudocode of the PSO.
\begin{algorithm}
\caption{Particle Swarm Optimization}
\label{algorithm:pso}
\small
\begin{algorithmic}[1]
\State $pop \gets generatepopulaion()$
\State $p \gets pop$ \Comment{set of best known solutions}
\State $best \gets getbestsolution(pop)$
\For{$k \in \{1 .. pop\}$}
	\State $v_k \gets U(-|b_{upper} - b_{lower}|, |b_{upper} - b_{lower}|)$
\EndFor
\State $i \gets 1$
\While{$j \leq iter$ $\And$ \ non stopping condition}
    \For{$i \in \{1 .. populationsize\}$}
	\State $x \gets pop_i$
    	\For{$f \in \{1..number\ of\ facilities\}$}
	    	\State $r_p \gets U(0,1)$
	    	\State $r_g \gets U(0,1)$
	    	\State $v_{i,f} \gets \omega v_{i,f} + \phi_p r_p (p_{i,f} - x_{f}) + \phi_g r_g (best_f - x_{f})$
	    \EndFor
	\State $x \gets x + v_i$
	\If{$fitness(x) < fitness(p_i)$} 
	    	\State $p_i \gets x$ 
		\If{$fitness(p_i) < fitness(best)$} 
	    		\State $best \gets p_i$
		\EndIf
	\EndIf
    \EndFor
    \State $i \gets i + 1$
\EndWhile
\State \Return{$best$}
\end{algorithmic}
\end{algorithm}
This swarm-based metaheuristic tries to continuously improve the whole population. In each iteration, the algorithm moves the solutions (modify them), which it calls particles, into more promising areas of the solution space. Each solution uses information from neighbouring solutions to find these areas.

Simulated Annealing was proposed by~\cite{SA}.
Its pseudocode is provided in Algorithm~\ref{algorithm:sa}.
SA employs a cooling strategy to estimate the current temperature $t$. This temperature is used by the potential function to accept the new solution (after shake) even if it does not improve to the best solution found so far. This behaviour allows controlling the exploration of the algorithm during its execution.
This algorithm has been used in different location problems~\citep{AROSTEGUI2006742,YU2015184,zarandi2015empirical}.

\begin{algorithm}
\caption{Simulated Annealing}
\label{algorithm:sa}
\small
\begin{algorithmic}[1]
\State $x \gets generation()$
\State $x \gets localsearch(x)$
\State $i \gets 1$
\While{$i \leq iter$ $\And$ \ non stopping condition}
	\State $t \gets cooling(t_0, i)$
	\State $k \gets next(i, iter)$
	\State $x^\prime \gets shake(x, k)$
	\If{$fitness(x^\prime) < fitness(x)$}
	    \State $x \gets x^\prime$
	\Else
	    \If{$random(0,1) < exp(-\Delta E / kt)$}
	    \State $x \gets x^\prime$
	    \EndIf
	\EndIf
	\State $i \gets i + 1$
\EndWhile
\State \Return{x}
\end{algorithmic}
\end{algorithm}

Variable Neighbourhood Search was introduced by~\cite{VNS}.
In this article we have based ourselves on the version used in~\cite{Drezner2015}.
The pseudocode is in Algorithm~\ref{algorithm:vns}.
This algorithm is based on the concept of neighbourhood. Each solution has a defined neighbourhood, i.e., a set of solutions with closest facilities to it. The current solution $x$ is modified according to these neighbourhoods ($next()$ indicates the number of modifications) and improved by a local search. In this version, a number $K$ of consecutive non-improvements is allowed before finishing the algorithm.
Versions of VNS have been used in localization problems as the \textit{p}-median problem reaching very good performance~\citep{Avella2012,Mladenovic2007,Reese2006}.

\begin{algorithm}
\caption{Variable Neighbourhood Search}
\label{algorithm:vns}
\small
\begin{algorithmic}[1]
\State $x \gets generation()$
\State $x \gets localsearch(x)$
\State $restart \gets true$
\While{$restart$ $\And$ \ non stopping condition}
	\State $restart \gets false$
	\State $j \gets 1$
	\While{$\neg restart$ $\And$ $\ j \leq K$}
	\State $i \gets 1$
		\While{$\neg restart$ $\And$ $\ i \leq k_{max}$}
			\State $k \gets next(i, k_{max})$
			\State $x^\prime \gets shake(x, k)$
			\State $x^\prime \gets localsearch2(x^\prime)$
			\State $x \gets acceptation(x, x^\prime)$
			\If{$x = x^\prime$} $\ restart \gets true$ \Else $\ restart \gets false$ \EndIf
			\State $i \gets i + 1$
		\EndWhile
		\State $j \gets j + 1$
	\EndWhile
\EndWhile
\State \Return{x}
\end{algorithmic}
\end{algorithm}

With this algorithms, we want to review the impact that local searches and shaking process, commonly used in trajectory algorithms, have on this type of problem.
Furthermore, thanks to GA and PSO, we want to check if a population-based strategy return better results than a trajectory-based one.
Besides, each of these algorithms has multiple choices of parameters and operators.
We have implemented several alternatives for each of them to give a broad view.
Among the possible parameter setting, we selected the best configuration to solve this problem, as we will see in Section~\ref{sec:irace}.
Next, we are going to give a brief description of each parameter as well as some abbreviations that we will use in the experimental section.

\textbf{General parameters}
    \begin{itemize}
        \item Number of iterations (iter). The total number of iterations of the algorithms can be calculated according to one of the following equations: $N \cdot p / 5$ (\texttt{Np/5}), $100 \cdot p$ (\texttt{100p}), $max(2 \cdot N, 100)$ (\texttt{2N100}), and one million of iterations (\texttt{1M}). Where $N$ is the number of customers and $p$ the number of facilities to select. These options have been extracted from various scientific articles.
        
        \item Domain model. We use two different models to assign a set of nearby facilities, of size $d$, to each facility (this model is used in shaking process): The facilities with the closest Euclidean distance (\texttt{NEAR}) and the closest by quadrants (\texttt{QUAD}). The latter model is calculated as follows: For each facility, the geographic space is divided into four quadrants using their coordinates (latitude and longitude). Then iteratively, we select the nearest point in each quadrant (if there are no more points, skip the quadrant) until we have picked $d$ facilities. The \texttt{QUAD} model provides more uniform coverage of the geographical space, at the cost of allowing more remote facilities. Figure~\ref{fig:neighbourhood-model} shows an example of these two models.
        
        \begin{figure}
        \centering
        \includegraphics[width=\linewidth]{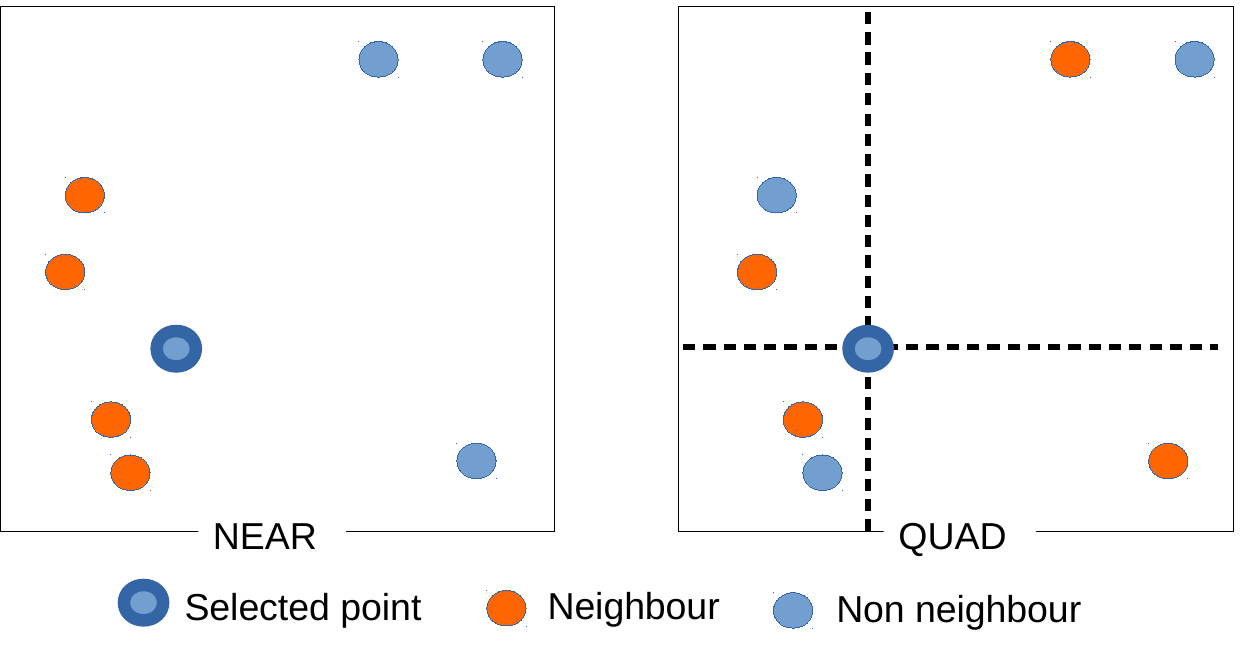}
        \vspace{-2em}
        \caption{Graphic example of the two domain models, $d=4$.}
        \label{fig:neighbourhood-model}
        \vspace*{-1em}
        \end{figure}
        
    \end{itemize}
        
 \textbf{Specific parameters of trajectory-based algorithms (ILS, SA, and VNS)}
    \begin{itemize}
        \item Local search. We can apply a local search procedure to the initial solution (localsearch) and in each iteration of the algorithm (localsearch2). The possible local search algorithms are: Fast Interchange (\texttt{FI})~\citep{doi:10.1080/03155986.1983.11731889}, Improved Alternate Approach (\texttt{IALT})~\citep{Drezner2016} with the parameter \texttt{Laux}, and Improvement Algorithm (\texttt{IMP})~\citep{Drezner2015} with the parameter \texttt{IMPparam}.
        \item Initial solution. We have four procedures to generate the initial solution: Random (\texttt{RAND}), the constructive algorithm proposed in~\cite{Drezner2015} (\texttt{START}), and the best of 100 random solutions (\texttt{100RAND}). 
        \item Shake. We use three different permutation algorithms to change $k$ facilities of a solution: Either we exchange the facilities with randomly chosen other ones in the set of close facilities defined in the domain model (\texttt{CLOSE}), we select the new facilities at random from all (\texttt{RAND}), or we do not perform any change of facilities to the solution (\texttt{NONE}).
    \end{itemize}
\textbf{Specific parameter of population-based algorithms (GA and PSO)}
    \begin{itemize}
        \item Population size: Number of individuals in the population. It is between 8 and 30.
    \end{itemize}

\textbf{GA parameters}
    \begin{itemize}
        \item $\lambda$. The number of new individuals to generate in each iteration of the algorithm. It can take values between 1 and the population size.
        \item Selection. To generate a new individual, the algorithm selects two solutions from the population. This selection can be random (\texttt{RAND}), selecting the two solutions with the best fitness value (\texttt{BETTER}), or the two with the worst fitness values (\texttt{WORSE}).
        \item Crossover. This operator combines two solutions (parents) to generate a new solution (offspring). It can be one of the following strategies: the merging process used in the GA described in~\citet{Drezner2015} (\texttt{MERGING}, one-point crossover (\texttt{1POINT}), the proposed crossing operator in~\citet{COLMENAR201888} (\texttt{CUPCAP}), or a copy of one of the parent solutions randomly selected (\texttt{1RANDPARENT}).
        \item Mutation. We use the same operations, \texttt{CLOSE} and \texttt{RAND}, as we do in the shaking process of the trajectory-based algorithms. We apply this operator with a probability in the interval $[0,1]$.
        \item Replacement. We can perform the replacement in the population according to one of these strategies: ($\mu, \lambda$) or ($\mu + \lambda$).
    \end{itemize}

\textbf{ILS parameters}
    \begin{itemize}
        \item Number of perturbations (npert). Between 1 and 20.
    \end{itemize}
    
\textbf{SA parameters}
    \begin{itemize}
        \item Next solution. We use two strategies to select how many facilities have to be shaked: The same value as the iteration index $i$ (\texttt{SEQ}) or the one used in the distribution based variable neighbourhood search proposed in~\cite{Drezner2015} (\texttt{DVNS}). The latter defines a probability distribution to avoid following the same pattern (sequential case) every time.
        \item Temperature ($t_0$). It is between 1 and 100.
        \item Cooling strategy. The strategy can be lineal (\texttt{LIN}), exponential (\texttt{EXP}), or the iteration index as the new temperature (\texttt{NONE}).
    \end{itemize}

\textbf{PSO parameters}
    \begin{itemize}
        \item $\omega$. Inertia parameter, in the interval $[0, 1]$.
        \item $\phi_p$. Cognitive parameter, in the interval $[0, 1]$.
        \item $\phi_g$. Social parameter, in the interval $[0, 1]$.
    \end{itemize}
    
\textbf{VNS parameters}
    \begin{itemize}
        \item Next solution. The same as in the SA.
        \item Exploration size (k$_{max}$). Number of neighbours that will be explored from the same solution between $[1, 50]$.
        \item Acceptance strategy. We have three options: Elitist, walk, or elitist but allowing to choose worse solutions according to a certain probability (acceptProb) between $[0, 1]$.
        \item Number of consecutive non-optimal solutions allowed (K). Between 1 and 100
    \end{itemize}

\section{Hardware platform}
\label{sec:hw}
The computation platform used in this work is composed of a cluster of 144 cores, equipped with three Intel Xeon CPU (E5-2670 v3) at 2.30$\,$GHz and 64$\,$GB memory.
The cluster is managed by HTCondor 8.2.7, which allows us to perform parallel independent executions to reduce the overall execution time. We have carried out 30 runs of each algorithm to statistically compare the results obtained.
The algorithms were run for 60 CPU seconds and reports the best solution found in each of the runs. The implementation was done in C\texttt{++} compiled with flags \texttt{-std=c++11} and \texttt{-O3}, the source code can be found in \url{https://github.com/cintrano/p-median}.

\section{Experimental Results}
\label{sec:experimentation}
In this section, after calculating the best parameter settings of each algorithm, we will study: 
(i) which algorithm works better to solve the problem,
(ii) which model is better adapted to the reality of the city, and 
(iii) how our system can assist to the municipalities to improve the expansion of their bicycle-sharing systems.

To carry out these studies, we took into account different data sets for the distances and weights (as we explained before in Section~\ref{sec:problem}) and we used the same number of stations $p = 23$ that are currently placed in Malaga.
The following are the results for each of the studies listed above.

\subsection{Parameter Setting}
\label{sec:irace}
Before starting with the results, we study which is the best parameter configuration that we found  for each algorithm.
The five algorithms have multiple variants in terms of operators and parameter settings.
Since manual multi-factorial complete experimental design is very time consuming, we use an automatic algorithm configuration stage through the use of the iterated racing procedure implemented by the irace package~\citep{LOPEZIBANEZ201643}.
This has the effect of producing a statistically best parameterised algorithm for our experiments. 
We included several components and parameters to give an overview of the algorithms.
The parameters were described earlier in Section~\ref{sec:algorithms}.
Table~\ref{table:tunning-final} shows the final parameters selected by irace for each algorithm.

\begin{table}[!b]
\footnotesize
\centering
\setlength\tabcolsep{3pt} 
\caption{Best parameter configuration found by irace.}{
\begin{tabular}{l|ccccc}
  \toprule
Parameter	&	GA	&	ILS	&	PSO	&	SA	&	VNS	\\
\midrule											
iter	&	\texttt{Np/5}	&	\texttt{1M}	&	\texttt{2N100}	&	\texttt{2N100}	&	\texttt{1M}	\\
localsearch	&	---	&	\texttt{IALT}	&	---	&	\texttt{IALT}	&	\texttt{IMP}	\\
Laux1	&	---	&	9	&	---	&	16	&	---	\\
IMPparam	&	---	&	---	&	---	&	---	&	2 \\	
domain m.	&	\texttt{QUAD}	&	\texttt{NEAR}	&	---	&	\texttt{QUAD}	&	\texttt{NEAR}	\\
d	&	20	&	29	&	---	&	34	&	49	\\
generation	&	\texttt{RAND}	&	\texttt{100RAND}	&	\texttt{RAND}	&	\texttt{100RAND}	&	\texttt{100RAND}	\\
shake	&	---	&	\texttt{CLOSE}	&	---	&	\texttt{RAND}	&	\texttt{CLOSE}	\\
npert	&	---	&	6	&	---	&	---	&	---	\\
next	&	---	&	---	&	---	&	\texttt{SEQ}	&	\texttt{SEQ}	\\
$t_0$	&	---	&	---	&	---	&	4.45	&	---	\\
cooling	&	---	&	---	&	---	&	\texttt{EXP}	&	---	\\
coolingOpt	&	---	&	---	&	---	&	0.39	&	---	\\
localsearch2	&	---	&	---	&	---	&	---	&	\texttt{IALT}	\\
Laux2	&	---	&	---	&	---	&	---	&	16	\\
k$_{max}$	&	---	&	---	&	---	&	---	&	5	\\
K	&	---	&	---	&	---	&	---	&	57	\\
accept	&	---	&	---	&	---	&	---	&	\texttt{ELITIST}	\\
population	&	14	&	---	&	29	&	---	&	---	\\
$\lambda$	&	10	&	---	&	---	&	---	&	---	\\
selection	&	\texttt{BETTERS}	&	---	&	---	&	---	&	---	\\
crossover	&	\texttt{CUPCAP}	&	---	&	---	&	---	&	---	\\
mutation	&	\texttt{RAND}	&	---	&	---	&	---	&	---	\\
mut. prob.	&	0.91	&	---	&	---	&	---	&	---	\\
replacement	&	$(\mu,\lambda)$	&	---	&	---	&	---	&	---	\\
$\omega$	&	---	&	---	&	0.0001	&	---	&	---	\\
$\phi_p$	&	---	&	---	&	0.53	&	---	&	---	\\
$\phi_g$	&	---	&	---	&	0.69	&	---	&	---	\\
  \bottomrule
\end{tabular}}
\label{table:tunning-final}
\end{table}

\begin{figure*}[t]
	\centering
    \includegraphics[width=\linewidth]{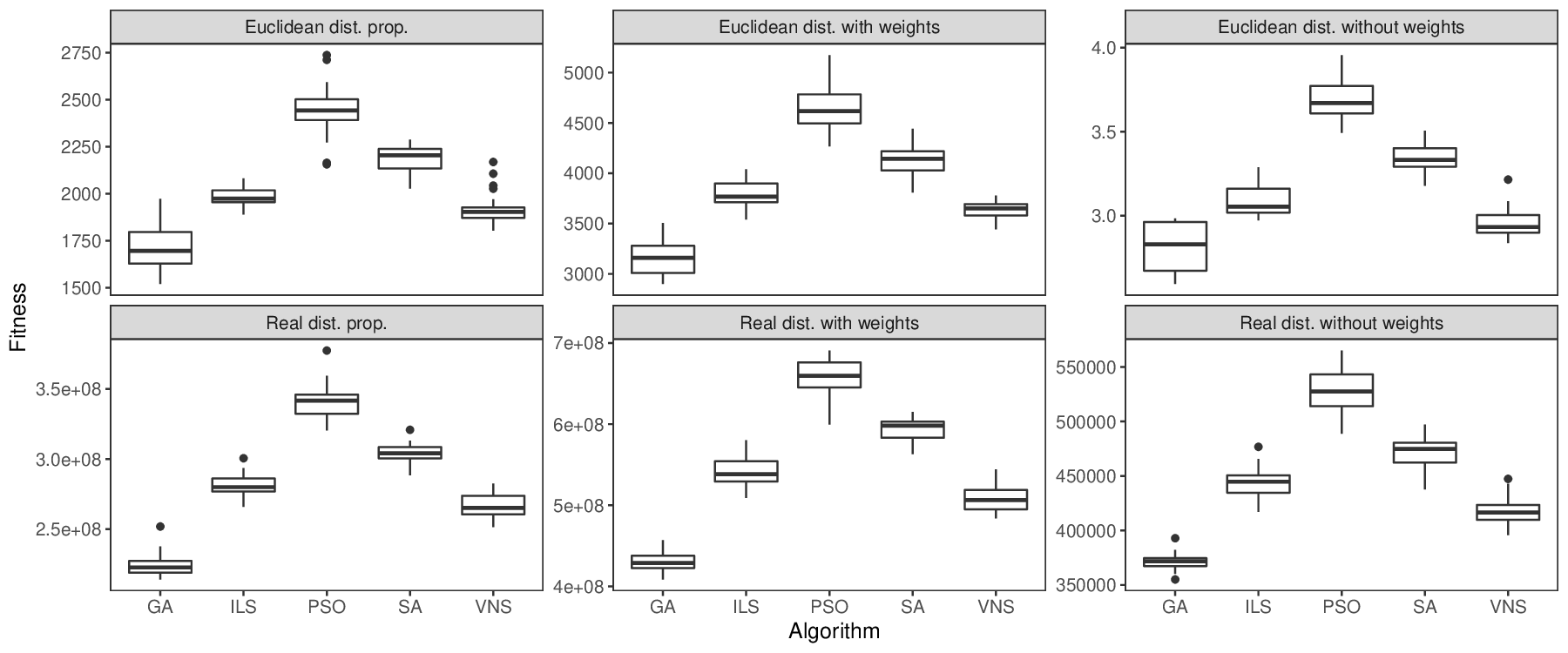}
    \vspace{-2em}
    \caption{Fitness value obtained in each of the scenarios for each algorithm.}
    \label{fig:imp-box-algo}
\end{figure*}

\subsection{Algorithm Comparison}
First, we will analyse the fitness values obtained by our optimisation algorithms in each scenario (type of distances and weights used).
Figure~\ref{fig:imp-box-algo} shows for each of our realistic scenarios the fitness values in the 30 independent executions.
In general, GA gets better fitness values than the other ones.
This highlights the power and versatility of these algorithms, a feature that has helped them gain popularity in part.
The second one is the VNS, this corroborates the popularity that this algorithm has when solving this type of localization problems.
GA shows a 5-16\% improvement over the second best algorithm.
The following positions correspond to ILS, SA, and finally, PSO, in all cases.
It is interesting to note that while one population algorithm obtains the best results, GA, another one obtains the worst results, PSO. The operators used in the GA may be better suited to solve this problem than those in the PSO. On the other hand, if we analyze the three trajectory-based algorithms between them, we see the importance of using a powerful local search to improve the performance of the algorithms.


In order to contrast these results, we are going to analyze whether the differences between the algorithms are statistically significant. We have performed a Wilcoxon Rank Sum Tests, with Bonferroni correction, between each pair of data sets. 
As we expected, there are differences between the different combinations of algorithms and weights, $\textit{p-value} < 0.01$ in all cases. 
In spite of this, and in view of the results, we can conclude that GA is, in general, the best option we have found to solve the problem of locating bicycle stations.

\begin{figure*}
	\centering
    \includegraphics[width=0.98\linewidth]{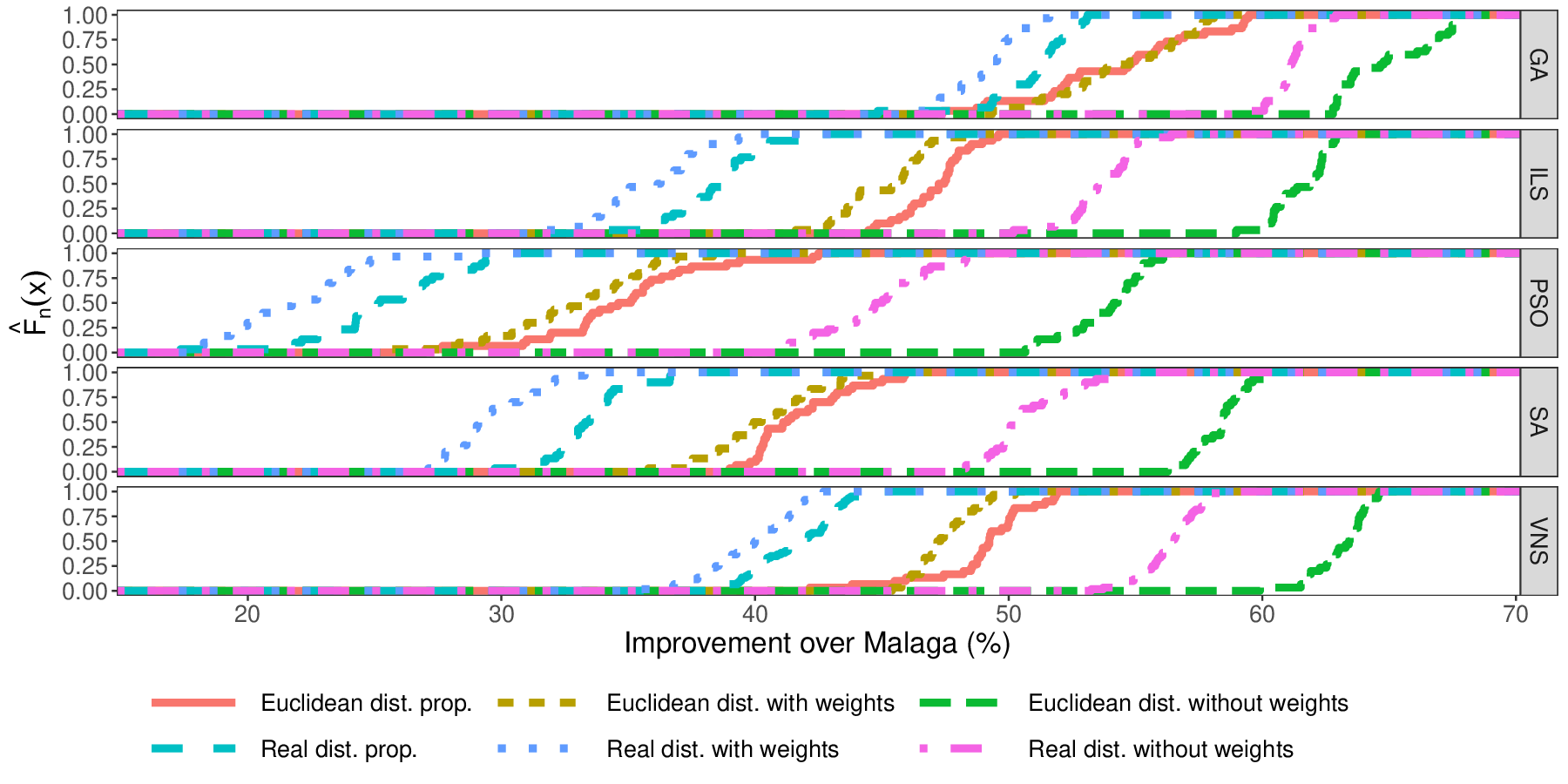}
    \vspace{-1em}
    \caption{Empirical cumulative distribution of the percentage of improvement of our solutions in each algorithm and scenario, compared to the current real location of bicycle stations in Malaga. We evaluate for each scenario the current solution of Malaga using the specific type of distance and weight in each case.}
    \label{fig:imp-algo}
\end{figure*}

\subsection{Improvement over the Current System}
After analysing the algorithms among themselves, we will compare them against the real solution in Malaga.
We compute the objective function for the actual location of the 23 stations in Malaga (evaluated according to each scenario).
Figure~\ref{fig:imp-algo} shows the empirical cumulative distribution of the percentage of improvement in the objective function value in each scenario and algorithm to Malaga.
In general, unweighted and Euclidean distance get the best rates of improvement over the actual city, around 55-65\% in distance reduction. 
If we do not take into account the most and least populated areas in the city, the stations will be set up spread across the city. With this behaviour, we reach more areas near the borders, reducing the distance to each district. 
However, the system in Malaga apparently takes into account the most densely populated area of the city, where it collects the highest number of bicycle stations. 
Even so, when we use information about the population, we obtain substantial improvements of about 50\% over the base scenario. 
An interesting aspect is that the type of distances or weights does not particularly affect the overall behaviour of the algorithms. In other words, there does not seem to be a unique key parameter to improve or give realism to the system, but a combination of them.

\begin{table*}[!ht]
\caption{Geographical distributions of the best solutions found by each algorithm in each of the six scenarios.}{
\centering
\setlength\tabcolsep{2pt} 
\begin{tabular}{c|c|c}
\toprule
 &                  Euclidean distance                         &                        Real distance                       \\
\midrule
\multicolumn{1}{c|}{\rotatebox[origin=l]{90}{$w_i = 1$}} & \includegraphics[height=0.21\textheight,width=0.47\textwidth]{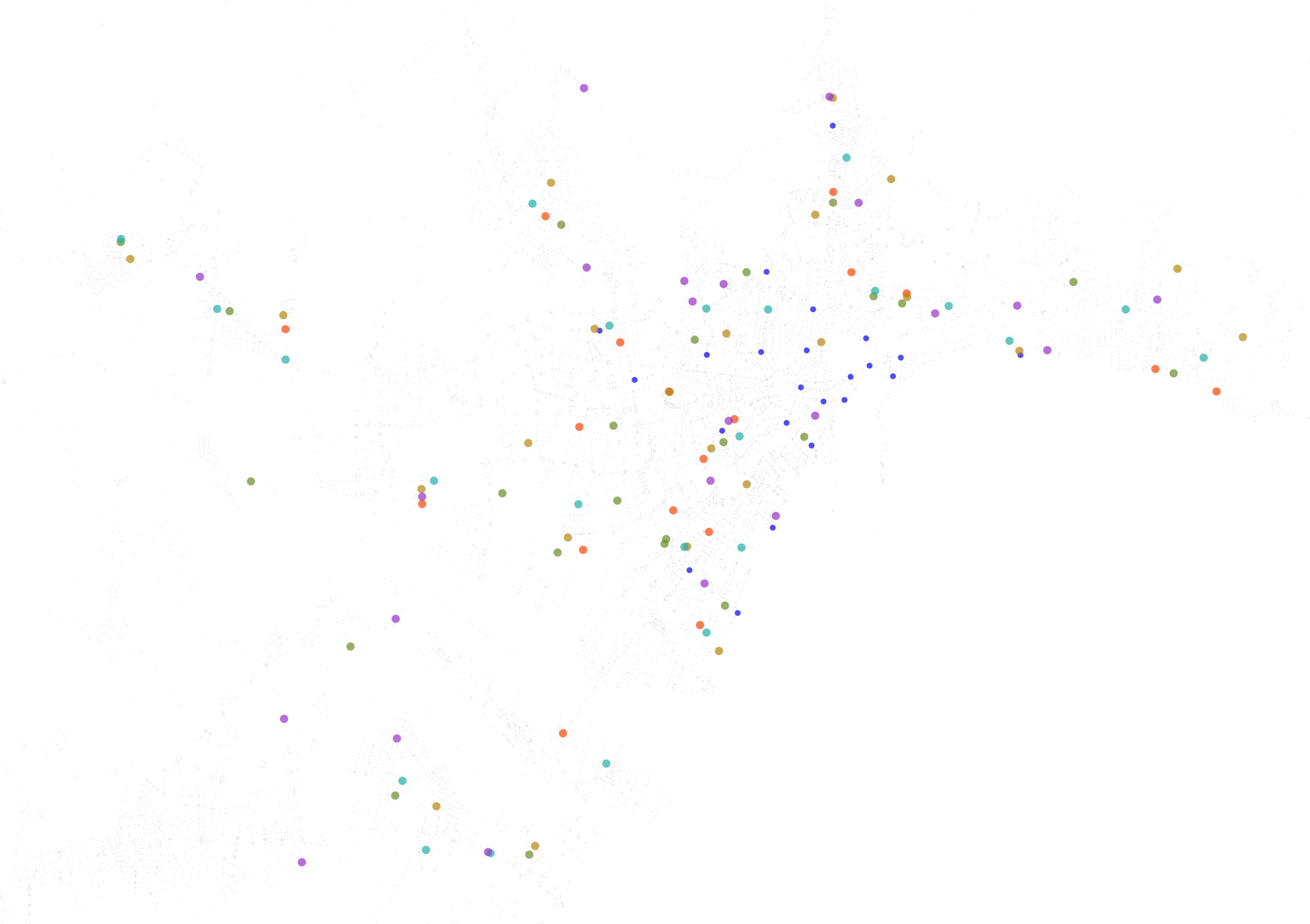} & \includegraphics[height=0.21\textheight,width=0.47\textwidth]{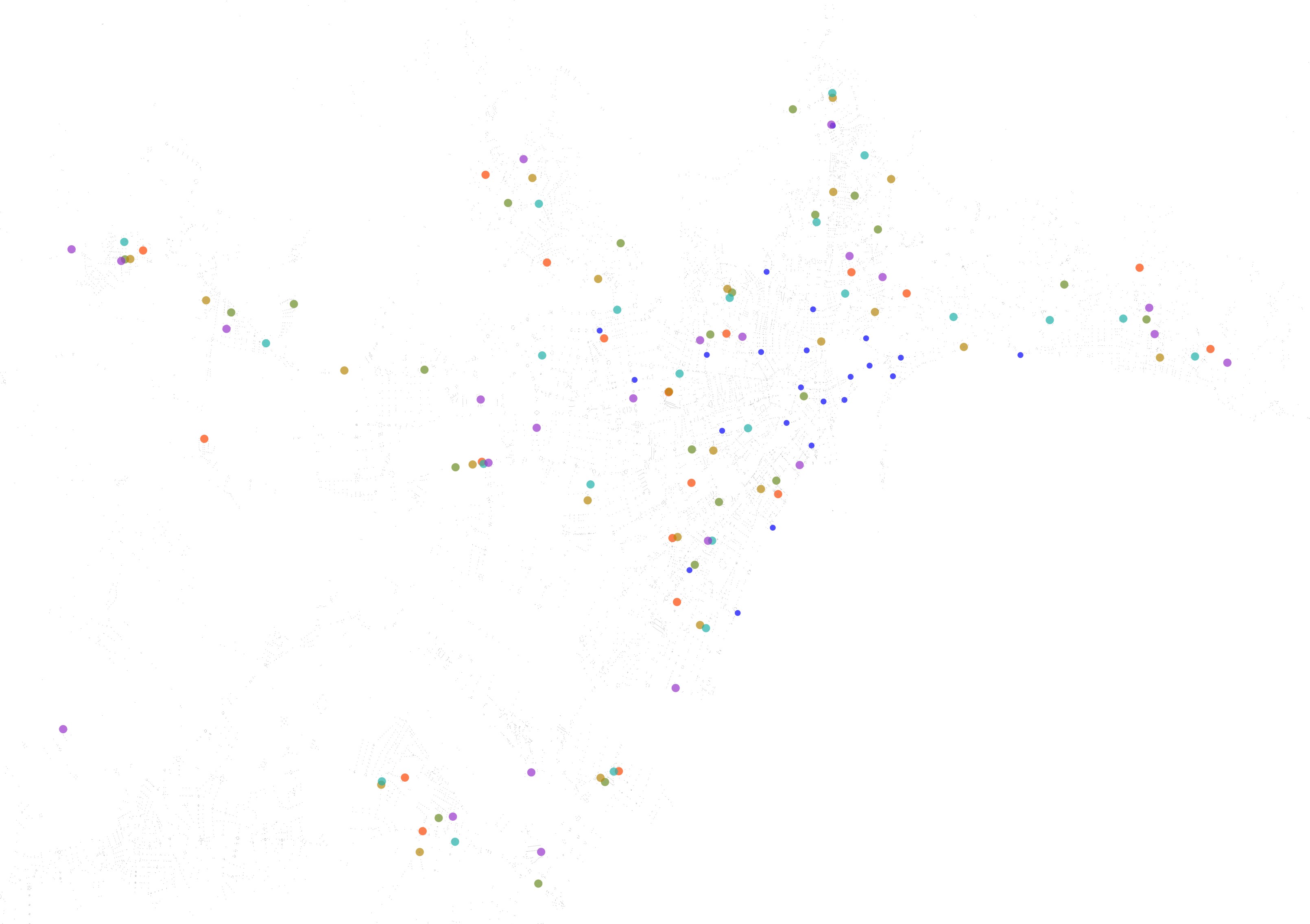}\\
\hline
\multicolumn{1}{c|}{\rotatebox[origin=l]{90}{$w_i = c_i$}} & \includegraphics[height=0.21\textheight,width=0.47\textwidth]{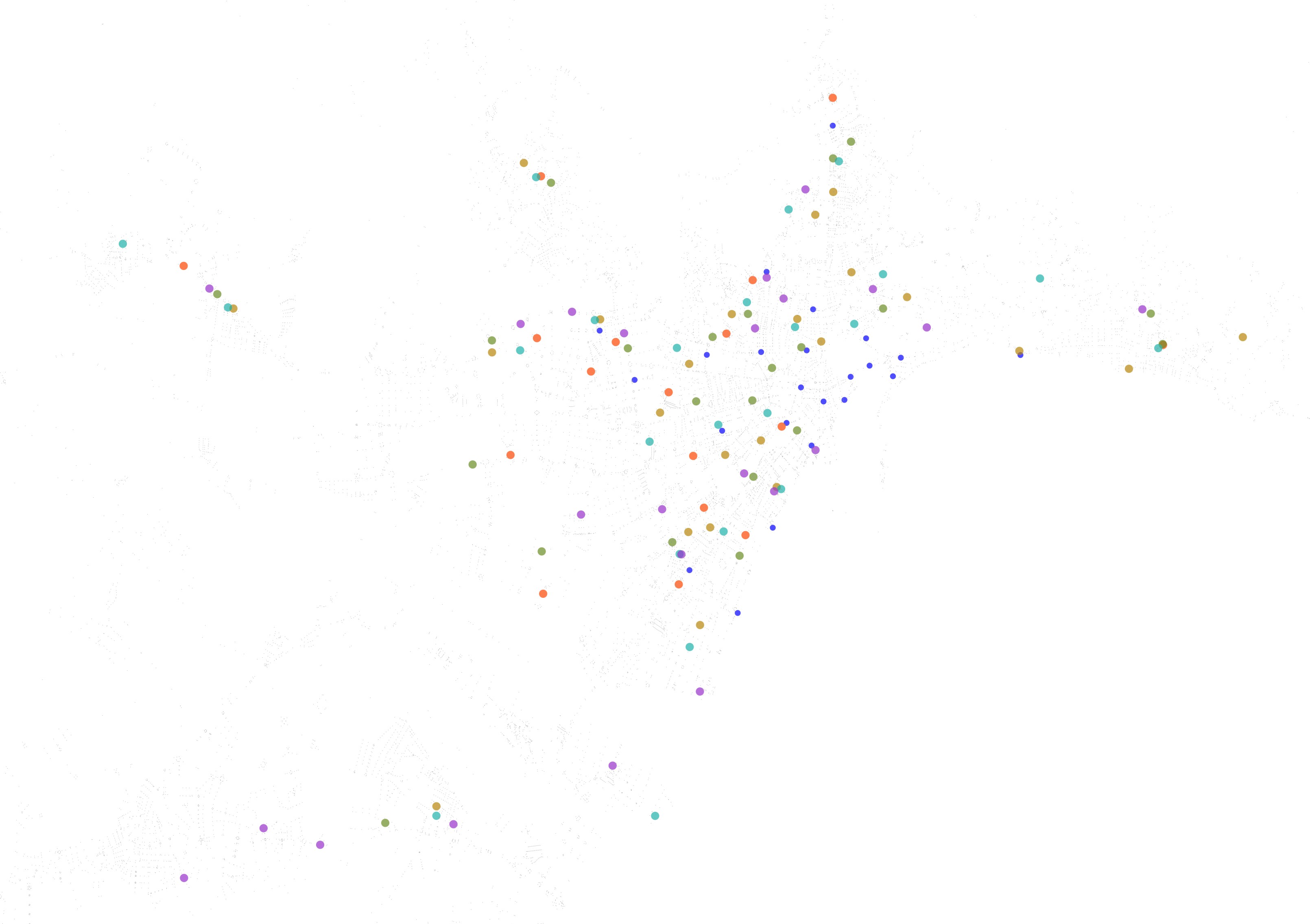} & \includegraphics[height=0.21\textheight,width=0.47\textwidth]{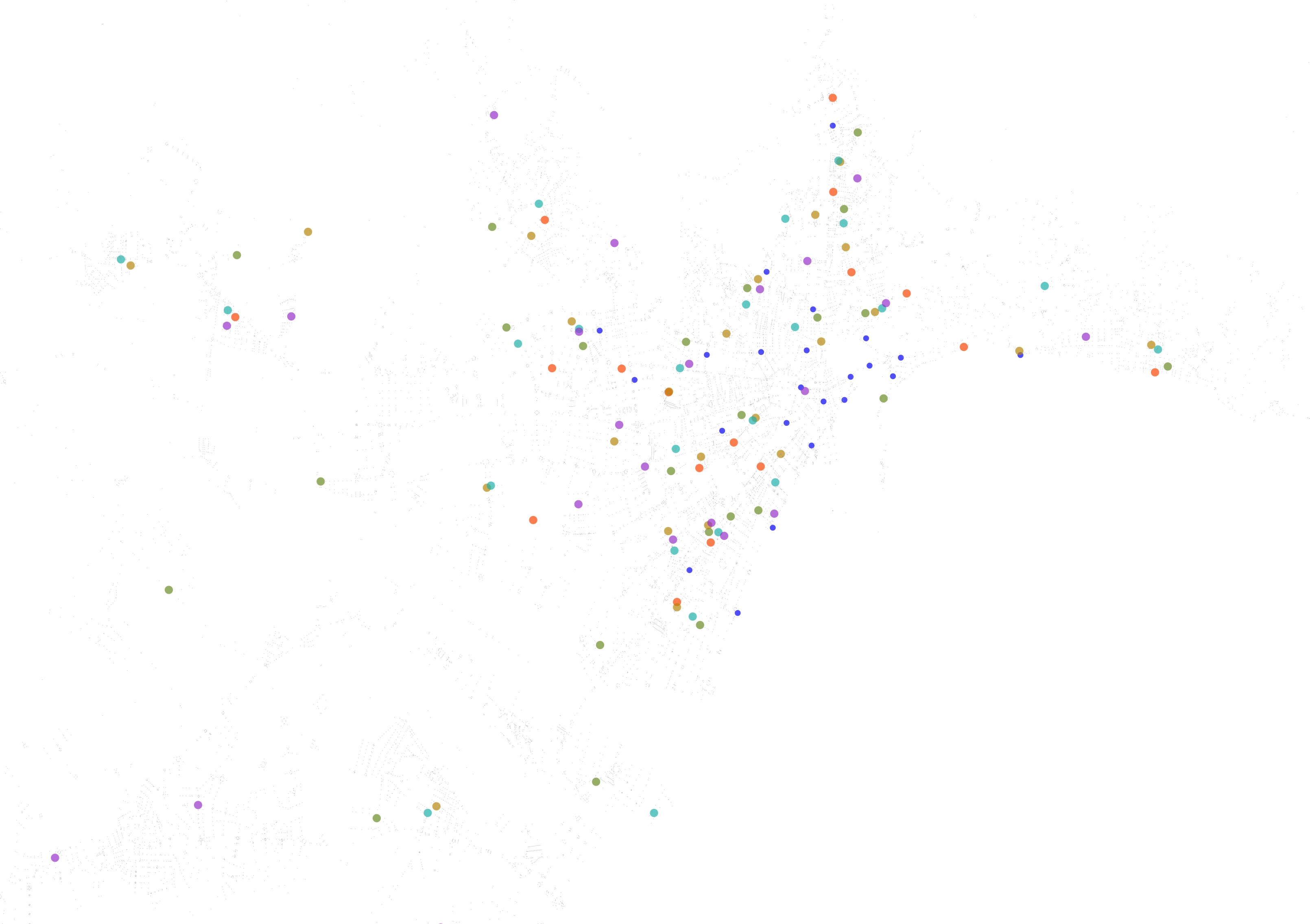}\\
\hline
\multicolumn{1}{c|}{\rotatebox[origin=l]{90}{$w_i = p_i$}} & \includegraphics[height=0.21\textheight,width=0.47\textwidth]{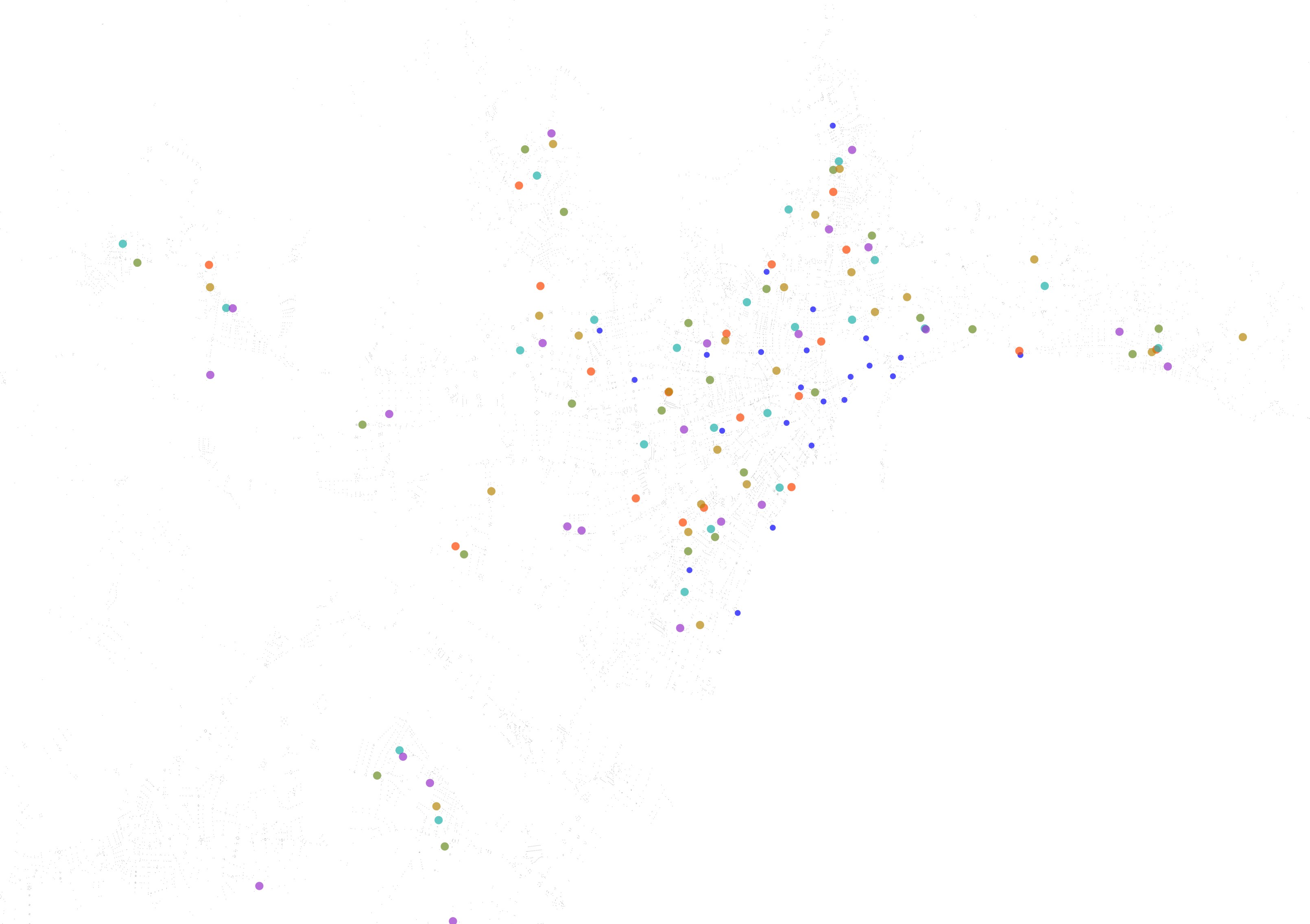} & \includegraphics[height=0.21\textheight,width=0.47\textwidth]{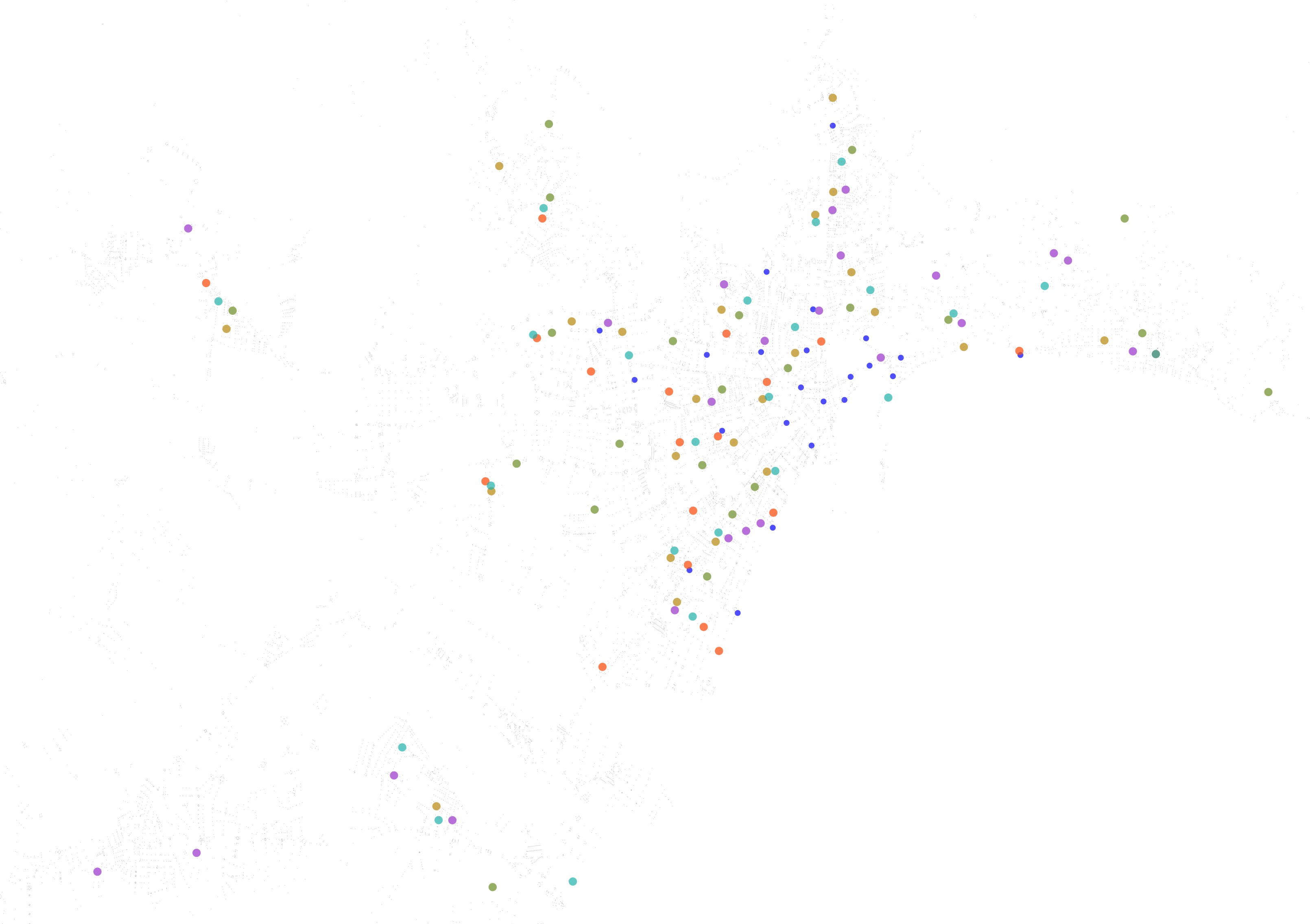}\\
\midrule
& \multicolumn{2}{l}{\small \begin{tikzpicture}
\node[mark size=3pt,color=blue] at (0,0) {\pgfuseplotmark{*}};
\end{tikzpicture} Current stations \begin{tikzpicture}
\node[mark size=3pt,color={rgb,255:red,32; green,178; blue,170}] at (0,0) {\pgfuseplotmark{*}};
\end{tikzpicture} GA \begin{tikzpicture}
\node[mark size=3pt,color={rgb,255:red,255; green,69; blue,0}] at (0,0) {\pgfuseplotmark{*}};
\end{tikzpicture} ILS \begin{tikzpicture}
\node[mark size=3pt,color={rgb,255:red,153; green,50; blue,204}] at (0,0) {\pgfuseplotmark{*}};
\end{tikzpicture} PSO \begin{tikzpicture}
\node[mark size=3pt,color={rgb,255:red,107; green,142; blue,35}] at (0,0) {\pgfuseplotmark{*}};
\end{tikzpicture} SA \begin{tikzpicture}
\node[mark size=3pt,color={rgb,255:red,184; green,134; blue,11}] at (0,0) {\pgfuseplotmark{*}};
\end{tikzpicture} VNS }  \\
\bottomrule
\end{tabular}}
\label{table:scenarios-maps}
\end{table*}

\subsection{Geographical Spread}
After discussing the quality of the results, we are going to study where in the city the found solutions place the bicycle stations. 
Table~\ref{table:scenarios-maps} shows the geographical positions of the stations in each instance of the problem.
For each of the combinations of distances and weights we show the solution with the lowest fitness value.
As we expected, the weighted versions allocate more stations in the central area of the city, which are the more populated ones.
But, they still offer more uniform coverage of the main neighbourhoods in the city than Malaga's system, so that each citizen has a nearby station to use the service.
In the weighted versions, there are usually no more than one or two stations in each of the neighbourhoods furthest from the city centre.
The quality of the service towards the citizens will be improved if the city council expends a small amount on infrastructure located in the outskirts.
On the other hand, the city's current solution uses many stations to cover the coastline.
However, our algorithms have shown that there are other avenues in the city that are more suitable to citizens.

\begin{table}[t]
\caption{Distance travelled per inhabitant to the nearest station in the different scenarios, evaluated as real distance and citizen weighting model ($w_i = c_i$). The minimum values in each column are marked in bold.}{
\centering
\scriptsize
\setlength\tabcolsep{2.5pt}
\begin{tabular}{c|c|c r r r r}
\toprule
Algo.	             & Dist.	                   & Weigh model & Min  &  Max 		& Mean 		& Median	\\
\midrule
\multirow{6}{*}{GA}	& \multirow{3}{*}{Eucl.}	& Uniform	& 929.51	& 1096.05	& 991.22	& 976.19 \\					
			&			& Citizens	& 738.55	& 890.20	& 810.11	& 815.00	\\
			&			& Demand	& 757.17	& 943.67	& 846.50	& 854.50	\\\cline{2-7}
			& \multirow{3}{*}{Real}	& Uniform	& 830.95	& 940.85	& 876.80	& 875.57 \\			
			&			& Citizens	& \textbf{719.16}	& \textbf{804.55}	& \textbf{756.48}	& 754.84	\\
			&			& Demand	& 730.77	& 848.97	& 760.73	& \textbf{754.62}	\\\hline
\multirow{6}{*}{ILS}	& \multirow{3}{*}{Eucl.}	& Uniform	& 940.23	& 1094.37	& 1018.23	& 1017.88	\\				
			&			& Citizens	& 905.33	& 1017.62	& 967.54	& 968.99	\\
			&			& Demand	& 896.95	& 1001.83	& 954.95	& 956.22	\\\cline{2-7}
			& \multirow{3}{*}{Real}	& Uniform	& 924.32	& 1048.65	& 989.06	& 991.78	\\		
			&			& Citizens	& 896.22	& 1021.60	& 950.88	& 945.79	\\
			&			& Demand	& 898.08	& 1007.24	& 947.37	& 945.20	\\\hline
\multirow{6}{*}{PSO}	& \multirow{3}{*}{Eucl.}	& Uniform	& 1049.01	& 1316.87	& 1192.22	& 1184.86 \\					
			&			& Citizens	& 1047.30	& 1315.17	& 1149.32	& 1147.55	\\
			&			& Demand	& 1032.34	& 1293.27	& 1150.80	& 1142.29	\\\cline{2-7}
			& \multirow{3}{*}{Real}	& Uniform	& 1053.65	& 1462.62	& 1198.15	& 1200.42 \\			
			&			& Citizens	& 1055.30	& 1216.14	& 1160.79	& 1161.21	\\
			&			& Demand	& 1056.60	& 1264.73	& 1143.84	& 1145.35	\\\hline
\multirow{6}{*}{SA}	& \multirow{3}{*}{Eucl.}	& Uniform	& 1009.83	& 1202.68	& 1102.89	& 1113.43	\\				
			&			& Citizens	& 935.82	& 1155.31	& 1033.12	& 1032.55	\\
			&			& Demand	& 961.25	& 1103.43	& 1043.84	& 1049.32	\\\cline{2-7}
			& \multirow{3}{*}{Real}	& Uniform	& 970.81	& 1221.67	& 1079.69	& 1071.86	\\		
			&			& Citizens	& 991.36	& 1083.02	& 1046.17	& 1052.98	\\
			&			& Demand	& 959.92	& 1074.92	& 1024.44	& 1023.37	\\\hline
\multirow{6}{*}{VNS}	& \multirow{3}{*}{Eucl.}	& Uniform	& 920.54	& 1056.32	& 993.77	& 995.12	\\				
			&			& Citizens	& 878.62	& 1001.12	& 933.26	& 924.58	\\
			&			& Demand	& 885.85	& 1056.88	& 933.54	& 931.91	\\\cline{2-7}
			& \multirow{3}{*}{Real}	& Uniform	& 898.52	& 1075.87	& 964.38	& 958.81	\\		
			&			& Citizens	& 851.56	& 958.11	& 894.28	& 891.61	\\
			&			& Demand	& 856.47	& 950.29	& 900.94	& 893.72	\\
\bottomrule
\end{tabular}}
\vspace{-1em}
\label{table:walking}
\end{table}

\subsection{Model Comparison}
Next, we are going to compare the different scenarios with each other. We want to find which of them suits the needs of the citizens. We have calculated for each solution the average distance that a person must travel from their home to the nearest station (real distance).
Table~\ref{table:walking} presents these distances in meters.
The shortest distances found were obtained by GA with the weights of the number of citizens. 
However, there is not much difference between the weight models $w=c$ and $w=p$ (even $w=p$ is better in some cases).

\begin{figure*}
	\centering
    \includegraphics[width=\linewidth]{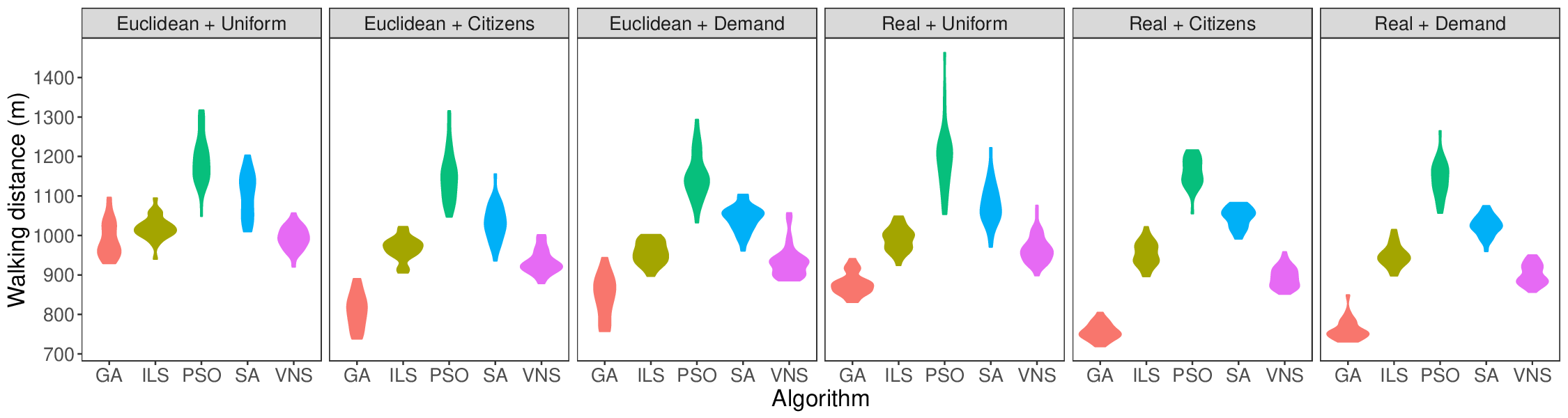}
    \vspace{-2em}
    \caption{Average distance (m) walked by the citizens to reach its nearest station. Each algorithm is shown in each combination of distances and weighting model.}
    \label{fig:violins}
\end{figure*}

We see in Figure~\ref{fig:violins} that the distributions for these weighting models are very similar, 
according to the different weighting models.
It is interesting to note that in algorithms that use local searches, in half the cases, two modes are observed instead of one.
This means that there are (at least) two local minima in the search space. Algorithms based on exploitation rather than exploration have more difficulties to get out of the local minima solutions.
On the other hand, GA presents very pronounced modes in the scenarios with real distances, which denotes a great convergence of the algorithm.
In general, all algorithms have a lower variability and get better results when using real distance instead of Euclidean.
All algorithms start the optimization process with solutions of similar quality, so the use of distances closer to reality makes the optimization process faster.

\begin{figure*}
	\centering
    \includegraphics[width=\linewidth]{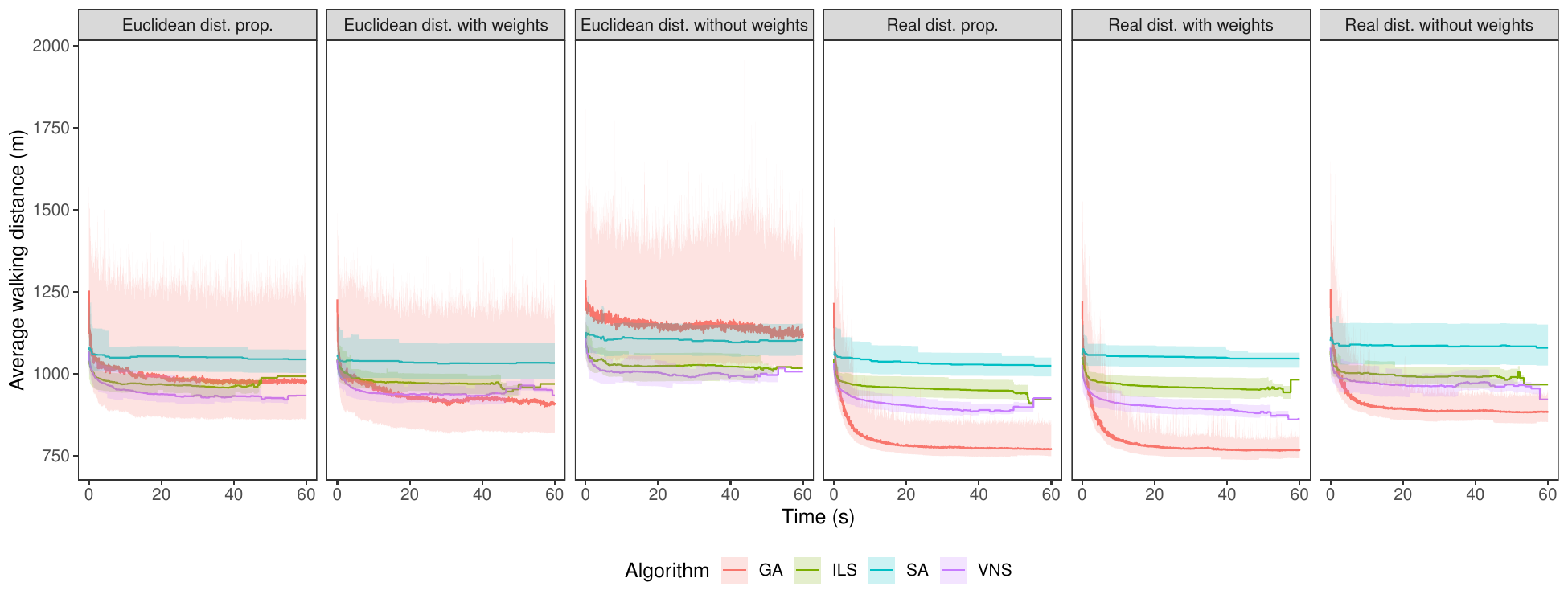}
    \vspace{-2.5em}
    \caption{Smooth function over the average distance walked by a citizen to its nearest station in the solution obtained in each iteration of the four betters algorithms. We decided to exclude PSO in order to clarify the figure, as its lowest value among all scenarios was 1,018.24\,m.}
    \label{fig:evolutions}
\end{figure*}

To confirm these results, we analyze the obtained solutions by the algorithms during their executions.
Figure~\ref{fig:evolutions} shows the mean distance walked by a citizen to the nearest station in the solution obtained during the running time.
We have represented both the average value and that obtained by $\pm 25\%$ of the solutions (smooth area) in each iteration.
We observe that the algorithms find a better solution faster when using real distances.
The quality of the solutions is more stable than in the other cases when we use a uniform weighting model. 
It is interesting to remark that GA has significant variability when we use Euclidean distance, obtaining even worse solutions than trajectory-based algorithms. On the other hand, when we calculate the solutions using the real distances, the competitive advantage that GA has become manifest. Notably, this difference is emphasized when we take into account information about the citizens.



\subsection{Increase in the Number of Stations}
\begin{figure*}[!htb]
	\centering
    \includegraphics[width=\linewidth]{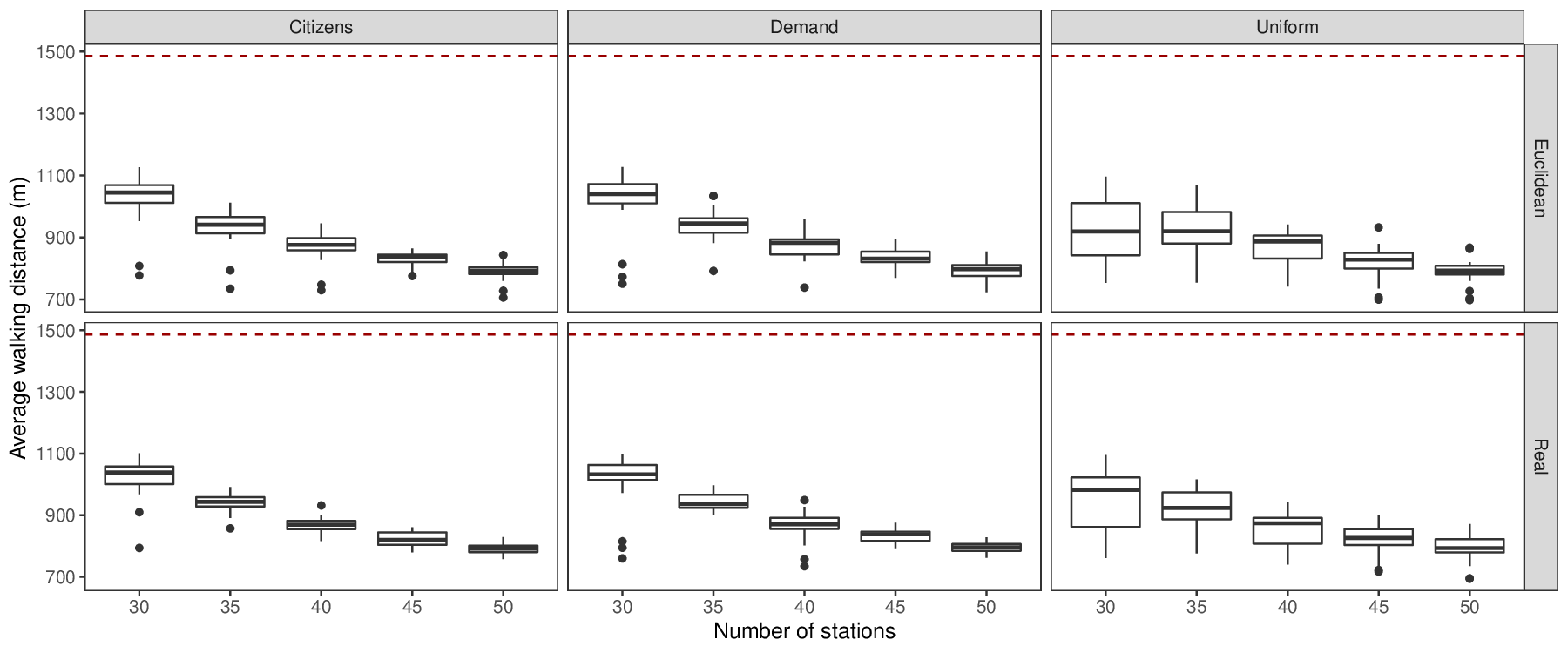}
    \vspace{-2em}
    \caption{Average distance (m) walked by the citizens to reach its nearest station when we increase the number of stations. The dashed red line indicates the average distance in the city's current system, 1485.97 m.}
    \label{fig:growth}
\end{figure*}

Finally, as a logistical issue, we are going to study how the city councils could improve the current bicycle shared system of the city. We added stations to the current solution in Malaga using the GA algorithm, that is the best algorithm in our experimentation. We have increased the 23 actual stations to reach 30, 35, 40, 45, and 50 stations and, thus, be able to verify the improvement obtained. 
Figure~\ref{fig:growth} shows the mean distance values obtained in each of our instances.
We get a decrease of a 33\% on average in the walking distance just by adding seven stations (smartly located). This is a great improvement for the citizens with a reduced cost in infrastructure. Furthermore, if we increase the number of stations to 50, the distance covered is reduced up to 53\%, which means that a citizen has a station only 738 meters away.
This expense in infrastructure allows the use of public bicycles to be a daily aspect for the population with the direct benefit of reducing the use of private vehicles and endorsing a healthy lifestyle.


\section{Related Work}
\label{sec:rw}
In bicycle sharing systems, there are multiple problems to address such as predicting the filling of stations~\citep{Singhvi2015}, the location-allocation of bicycles~\citep{Chen2018,Chira2014} in the stations, routes for users or the transfer of bicycles~\citep{Hu2014}, etc.
These are interesting problems based on the presence of an already installed system (including the associated infrastructure). We think that it is essential to apply intelligent techniques from the very moment of planning the final location of the different elements of the system.
There are also complete solutions that take into account multiple aspects of this kind of systems~\citep{Lin2013}.
However, they are very complex solutions that require large amounts of information that are not always available.

If we focus on the optimal location of the stations, we find approaches such as those proposed in~\cite{Chen2015}. In this paper, the authors use real data and machine learning techniques to find the best places to allocate the stations.
A similar approach to the above is used in~\cite{Liu2015a}, where they use New York City as case study, which has a large network of bicycle stations.
Furthermore, \cite{Park2017} present a comparison between two models for the station location problem.
The authors compare the \textit{p}-median and the MCLP. However, they do not give details about the algorithm that is used. \cite{Chen2015a} try to find the best locations so as to ensure the availability to collect/deposit the bicycles, taking into account the demand (at peak times) and the possible routes between stations made by users.
These works involve a demand on the use of the bicycle-sharing system.
However, they do not take into account potential users (citizens) who could use the system if it were closer to them.

Our work is the only one to use demographic data to bring the system closer to all users (which has not been taken into account in other works). In addition, the existing solutions use a custom formulation, making it difficult to compare against each other state-of-the-art solutions.
We formulate the problem of locating bicycle stations as a \textit{p}-median which allows us to use a solid base for our studies, as well as to enrich ourselves with all the advances made in the \textit{p}-median problem.

\section{Conclusions}
\label{sec:conclusions}
In this work, we remarked how solving the classic location problem \textit{p}-median we can solve the problem of where to place bicycle stations. We analysed the state-of-the-art and selected five algorithms commonly used to solve the \textit{p}-median and others NP-hard problem: GA, ILS, PSO, SA, and VNS. Each of them was tunned using irace over a set of benchmark instances. In addition, because we are solving a problem for the cities, we used several realistic data from the city of Malaga as an instance to test our work. The main conclusions of the work are the following:
\begin{itemize}
    \item GA has been the algorithm that returned the best results, opposite to the state-of-the-art.
    \item Compared to the assignment made by an expert of the city council, we obtain improvements of 50\% of quality when we apply metaheuristic techniques.
    \item Real distance allows us to find better locations for the bicycle stations, and number of inhabitants per neighbourhood enables us to accommodate the results to the citizen's needs. 
    \item Our proposal can be used to improve the city's current system adding stations, obtaining improvements between 33\% and 53\%.
\end{itemize}

As future work,  
we want to extend the applicability of the model in the real world, not only including information from the point of view of the client (location of neighbourhoods and demand for use by citizens) but to include data on the morphology of the city, types of roads, number of slots per station and the importance of certain points of interest (hospitals, schools, shopping centres, ...).
Another future work would be to expand our model with other parts of a commercial system, such as transporting bicycles from one station to another to balance the load, or studies of taxes and economic impact on citizens.
Besides, we would like to add even more operators to each of the algorithms and perform a more exhaustive analysis of the impact of the operators in the global performance.
Also, we want to apply the \textit{p}-median as the model for the electric car charging points allocation problem, because its growth makes it a necessary challenge for today's cities.

\section*{CRediT authorship contribution statement}
\textbf{C. Cintrano:} Conceptualization, Methodology, Software, Validation, Formal analysis, Data Curation, Writing - Original draft preparation, Visualization. \textbf{F. Chicano:} Investigation, Resources, Data Curation, Writing - Review Editing,Supervision, Project administration. \textbf{E. Alba:} Resources, Writing - Review Editing, Funding acquisition.

\section*{Acknowledgements}
This research was partially funded by the University of M\'alaga, Andaluc\'{\i}a Tech, the Spanish MINECO and FEDER projects: TIN2014-57341-R, TIN2016-81766-REDT, and TIN2017-88213-R.
C. Cintrano is supported by a FPI grant (BES-2015-074805) from Spanish MINECO.


\end{document}